# Biomorphic propulsion system diving thunniform robotic fish

*I.V. Mitin*[1,2], R.A. Korotaev[1,2], A.A. Ermolaev[2] and V.B. Kazantsev[1,2,3]

[1] Immanuel Kant Baltic Federal University, Kaliningrad, Russia

[2] National Research Lobachevsky State University of Nizhny Novgorod, Nizhny Novgorod, Russia

[3] Russian State Scientific Center for Robotics and Technical Cybernetics, St. Petersburg, Russia

\* Corresponding e-mail: illya.mitin@gmail.com

**Abstract**

A biomorphic propulsion systemfor underwater robotic fish is presented. The system is based on a combination of an elastic chord with a tail fin fixed on it. The tail fin is connected to servomotor by two symmetric movable thrusts simulating muscle contraction. The propulsion system provides oscillatory tail movement with controllable amplitude and frequency. Tail oscillations results in translational movement of the robotic fish implementing the thunniform principle of locomotion. The shape of the body of the robotic fish and the tail fin were designed using computational model simulating virtual body in an aquatic medium. A prototype of robotic fish device was constructed and tested in experimental conditions. Dependencies of fish velocity on the amplitude and frequency of tail oscillations were analyzed.

**Introduction**

Design of biomorphic robots has attracted great attention of researchers from different fields. Such robots intend to reproduce principles of movement of living creatures in nature. These principles have been optimized in adaptive interaction with nature during millions years of evolutions permitting creatures to survive. In technical terms the survival is defined by two obvious basic points. The first one concerns minimizing of energy consumption during movement. The second one is maximizing movement efficacy, e.g. speed, or distance travelled, or acceleration, in conditions with limited energetic resources. Another important point concerns way of interaction with natural environment. Specifically, movement should be implemented in the way minimizing possible perturbations of physical parameters of air or aquatic environment. Obviously, the design of creature like biomorphic machines represents very attractive engineering task [1-13].

The design of robotic fishes represents one of the most developed modern areas of bionic engineering. Many fish robot solutions have been proposed in recent decades. Following fish classifications robots can be also grouped on the type of locomotion implemented. The thunniform locomotion is based mostly on tail fin oscillatory movement while the whole body practically does not bend [1,2]. On the opposite anguilliform locomotion involves wave-like oscillations of the whole body [10]. The thunniform swimming is realized in tunas representing big ocean fishes capable to travel over long distances with sufficiently high speed. Following natural shape and geometry many different robotic solutions were proposed. Robot sizes differed from meters in length [3] to dozen of centimeters [4]. The propulsion systems are typically based on servomotor driving one tail fin [3]. Oscillatory tail fin movement yields robot translational motion. Robots are also supplied with different sensors and navigation systems providing autonomous diving in water pool.

Particular questions arisen in the biomorphic robot design include the choice of optimal parameters of the body shape, fin shape and size and also dynamical parameters, e.g. amplitude and frequency, of the propulsion system. It was recently shown that body flexibility and stiffness can significantly influence on swimming speed [4].Strategic aim is to approach to real animal efficiency in the robotic propulsion system performance. To solve such optimization task both computational simulations and physical experiments can be applied [2,4].

In this paper we present the design of a robotic fish with thunniform locomotion with biomorphic propulsion system. The robot is supplied with a flexible cord with oscillating tail fin. We investigated a computational model of robot movement in aquatic environment and performed physical experiments with robot swimming in a water pool analyzing the influence of dynamical parameters of the propulsion system on the robot efficiency.

**Robotic Fish Model**

Figure 1 Ashows a general computer model of our thunniform fish robot. The shape of the robot body was scaled from real geometry of tuna fish.Based on the photographs, a 3D model was built, then it was adjusted for the equipment that should be placed in the case. During the development, the NX computer-aided design system was used. The robot is equipped with control boards and a WI-FI module. Responsible for maneuvering is the dive-ascent system, swivel fins and the propulsion system itself. The robot is equipped with a camera, ultrasonic rangefinder and temperature sensor.The model was implemented in the form of robotic device that was used in physical experiments (see Fig. 1B). The propulsion system is located in the back part of the robot and shown in details in Fig. 2A. It consists of servomotor fixed on the robot body, flexible plate and tail fin.On both sides of the flexible plate, there are rods that deform it when the servo rotates. To give an additional degree of freedom, the tail fin is mounted on a spring-loaded hinge. The rods are sealed with silicone bellows (see Fig. 2B).

Tail oscillations are provided by two wire rods playing the roles of flexors and extensors mimicking muscular contractions. In the result robotic fish demonstrates thunniform movement when the tail oscillations are converted into translational motion of the whole robot. In our model the control system can change gradually the amplitude and the frequency of the tail oscillation.

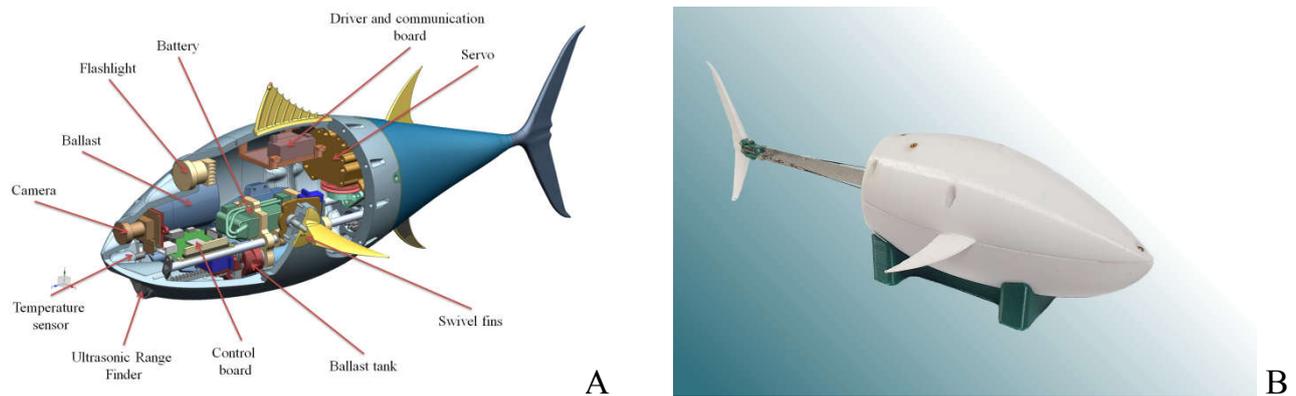

**Figure 1.** A. General view of the robot model. The body shape was copied from real tuna geometry. The location of basic components of control, navigation and sensor systems. The propulsion system is located in the back part of the robot driving the tail fin. B. Working prototype of thunniform robotic fish.

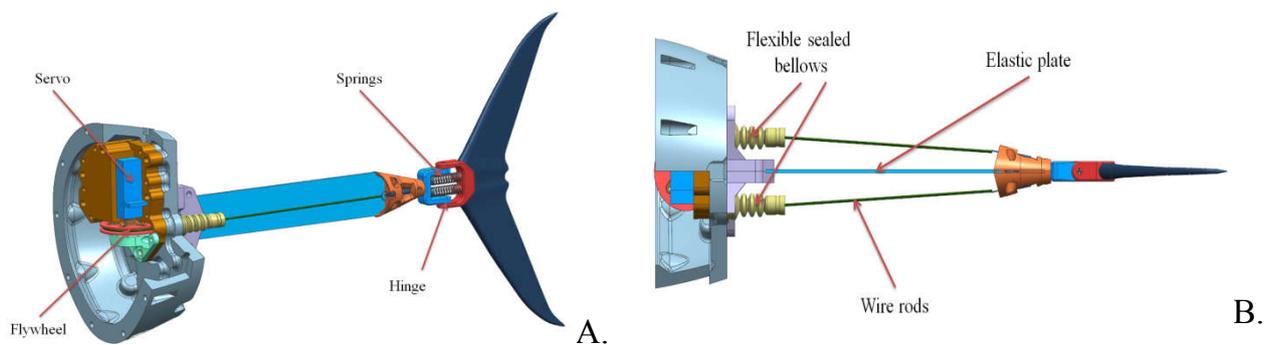

**Figure 2.** Robotic fish propulsion system. A. Tail fin is connected to a servomotor. B. Muscle like wire robots providing flexor/extensor movement of the tail fin mounted to an elastic rod.

In computer simulations we also tested the robot movement in an aquatic environment using ANSYS software. The main goal of optimizing the movement of a fish-like drone is to achieve the best possiblerelationship between tail geometry and power consumption of the drive, which is the basisdriving force. A prototype was developed to verify the computer modelbiomorphic mover, allowing to simulate the movement of fish.Calculationsinclude several stages: buildinga computer model from different computational modules presented inside the program,loading the studied geometry (3D model), building a computational grid on the surfacethe geometry under study, the problem of the initial conditions, directly carrying out the calculation andanalysis of the results obtained. Specifically, we looked on the hydrodynamicthe body resistance arising in the fluid flow using our 3D model. The results are shown in Figure 3.

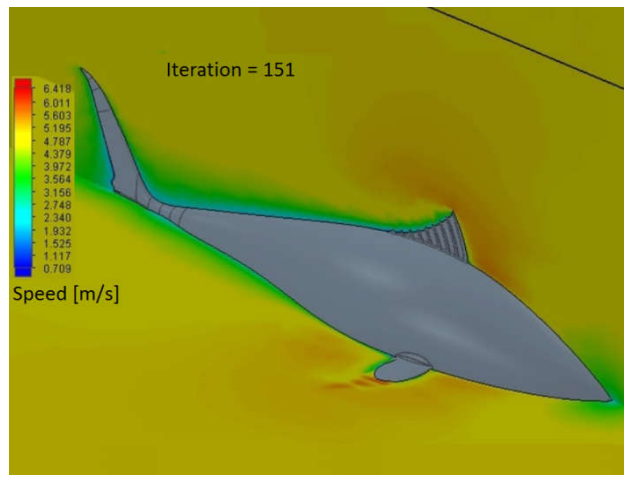

**Figure 3.** Simulation of body resistance profile of the 3D fish robot movement in an aquatic medium with constant flow. Color grade corresponds to values of the flow speed.

**Fish swimming experiments**

For physical experiments with the robotic fish device we developed the experimental setup shown in Figure 4.

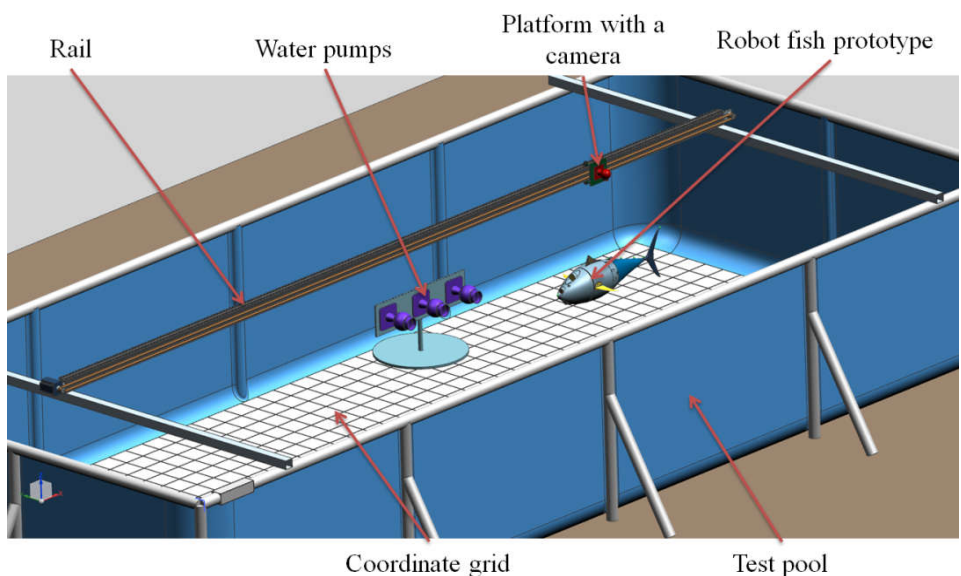

**Figure 4.** Experimental setup to monitor characteristics (speed, direction, mode of movement) of the robotic fish.

The setup is capable of registering the kinematics of swimming and disturbances of fluid flows. It consists of a carriage with a camera attached to it, which moves along rail guides. The camera captures the movements of the robot against the background of the coordinate grid, which then makes it possible to further digitize the movements.

A number of experiments have been carried out to test the swimming of a fish-like robot in a pool. The robot was placed in the pool (Fig.4), then a control command was sent from the computer via WI-FI. The control team started the movements of the detection unit with a given frequency and amplitude of tail oscillations. The robot's swimming was recorded on video. In different series of experiments, different initial parameters of the frequency (from 0.5 to 7 Hz) and amplitude (from 20 ° to 80 °) of the strokes were set.

The protocol of the experiments was the following. In the first part of the experiment, a series of launches of the robot was carried out at differentthe frequency and amplitude of the tail flaps. Before each launch, the controllers were set with parameters of the amplitude and speed of the servo that did not change duringseparate launch. The robot was moving in the water column, over a marked grid with a cell spacing of 100x100mm. While the robot was moving, filming was carried out on a video camera with further processingimages on the computer. A segment of the path was selected on which the robot's speed wasconstant (after starting for a while, it accelerates until it reaches a constant speed),the length of the distance traveled (according to the markings) was determined and the time it took for the robot to swim thispath. Then the speed of translational movement was calculated (not taking into account the rocking of the robotfrom the flapping of the tail). A number of starts were carried out with different control parameters andthe velocities are calculated for each resulting mode. Figure 5 illustrates swimming robot in the water pool.

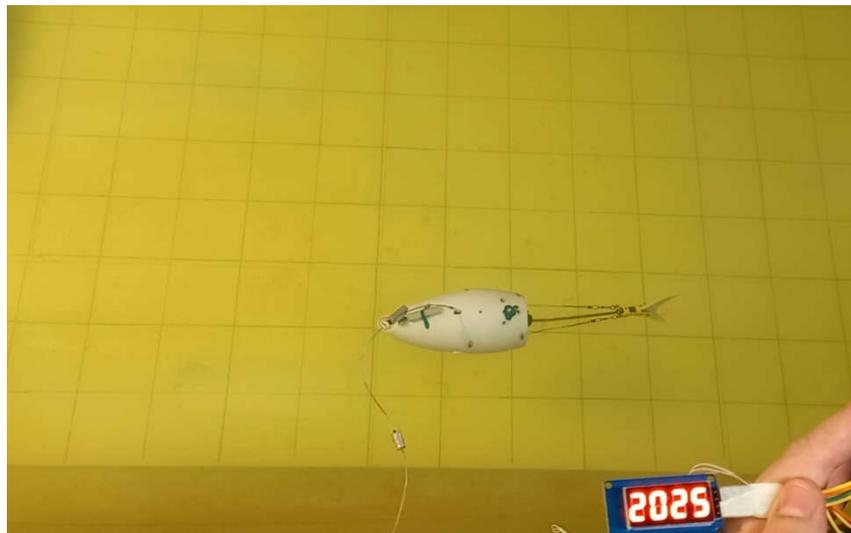

Figure 5. A picture of experimental swimming. Robotic fish relative to square grid.

Next, it was necessary to move from the control parameters of the servo to realthe parameters of the period and amplitude of the tail flaps. To do this, in the second part of the experiment, werepeated a series of servo starts with the same control parameters, but infixed position of the body with video recording against the background of the coordinate grid. Furthermore, the actual values of the amplitude and period of the tail oscillations were determined in the video materialfor each operating mode. The results illustrating dependences of actual robot speed on the tail oscillation amplitude and period are presented in figure 6.

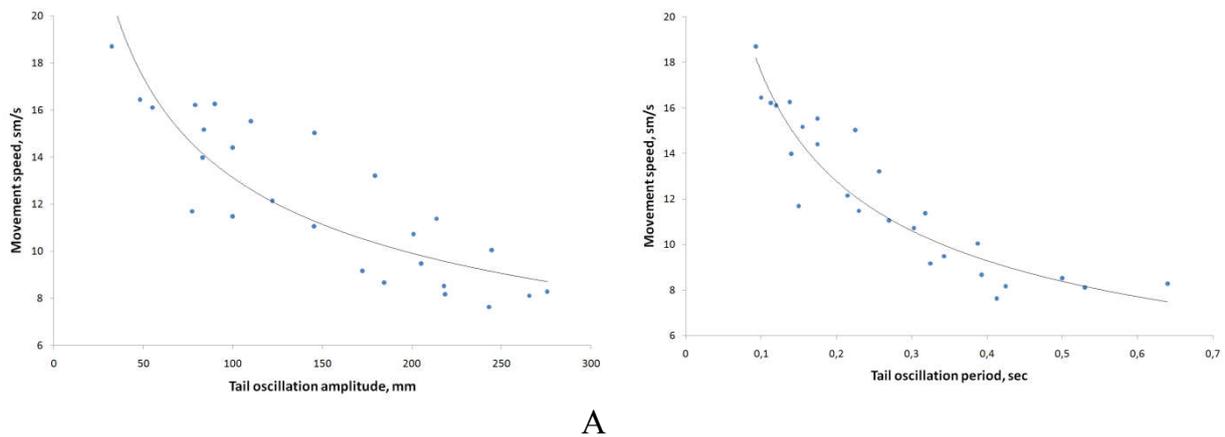

Figure 6. Results of experimental swimming of the thunniform robotic fish. A. The dependence of the movement speed on the tail oscillation amplitude. B. The dependence of the movement speed on the tail oscillation period.

We found that the highest speed of translational motion was observed with a small period of oscillations and their small amplitude, respectively. With an increase in the amplitude, the body started to sway around the longitudinal axis, which reduced the efficiency of the translational movement. With a different shape of the body, for example, as in fish with a flattened body elongated along the vertical axis or a large area of fins located in a vertical plane, this parasitic swaying would not occur. However, for the shape characteristic of tuna, flapping with high frequency and low amplitude showed the greatest efficiency.

**Conclusions**

We have developed thunniform swimming robot imitating tuna fish movement. We have proposed a novel propulsion system composed of elastic cord with flexor/extensor mechanical systems mimicking muscular contraction in living animals. This system provides oscillatory movement of tail fin that, in turn, generates translational motion of the robot. The shape and geometry of the robot body reproduces tuna parameters was first designed in computer model, next implemented in working robotic device. Series of experiments investigating how the robot kinematics depends on the dynamic parameters of the propulsion system were carried out. General conclusion was that for such animal shape and thunniform locomotion modes lower amplitude tail fin oscillations with relatively high frequencygenerate maximal speed.

We also note that experimental results with general tendencies in decreasing dependences (see, fox example, Figure 6) have quite high degree of variability. This is a result of problem complexity when 3D body dynamics in the hydrodynamical medium can acquire additional degrees of freedom competing with translational motion and consuming energy. To overcome these problems dynamical feedbacks should be further invented compensating such physical fluctuations. Note, that in recent studies such dynamical adjustments of the cord stiffness have revealed significant increase in the swimming performance [4]. We also plan to include an adaptive feedback into the control systems driving the tail oscillations that would compensate non-translational fluctuations and dynamically converge to an optimal swimming mode.

**Acknowledgements**

This work was supported by the Russian Science Foundation (grant No. 21-12-00246).